\documentclass[runningheads]{llncs}

\usepackage{eccv}

\usepackage{eccvabbrv}

\usepackage{graphicx}
\usepackage{booktabs}
\usepackage{caption}

\usepackage[accsupp]{axessibility}  % Improves PDF readability for those with disabilities.

\usepackage[pagebackref]{hyperref}
\usepackage{orcidlink}
\usepackage{multirow}
\usepackage{wrapfig}

\newcommand{\err}[1]{\,\scalebox{0.85}{$\pm\,#1$}}

\begin{document}

% ---------------------------------------------------------------
% TODO REVIEW: Replace with your title
\title{Synesthesia via Direct Latent Augmentation: \\Bypassing the Decode-Encode Loop for Cross-Modal Distillation} 
\titlerunning{Synesthesia via Direct Latent Augmentation}

\author{Cristian Sbrolli\inst{1}\orcidlink{0000-0001-6873-1967} \and
Nicolas Michel\inst{2}\orcidlink{0000-0002-6265-6748} \and
Matteo Matteucci\inst{1}\orcidlink{0000-0002-8306-6739} \and
Toshihiko Yamasaki\inst{2}\orcidlink{0000-0002-1784-2314}}

\authorrunning{C.~Sbrolli et al.}

\institute{Politecnico di Milano, Milan, Italy \and
University of Tokyo, Tokyo, Japan \\
\email{cristian.sbrolli@polimi.it}}

\maketitle
\begin{abstract}
While multimodal integration significantly improves computer vision models, deploying them incurs prohibitive inference costs and requires scarce, perfectly paired datasets. Recent methods address this data bottleneck by synthesizing missing modalities via generative AI, yet they introduce a severe inefficiency: the Decode-Encode Loop. Specifically, information-rich generative latents are decoded into noisy raw signals, forcing the downstream classifier to waste capacity re-encoding them. To bypass this bottleneck, we propose Direct Latent Augmentation (DLA), utilizing undecoded generative latents directly as privileged information. Furthermore, to transfer this dense knowledge to a purely visual student, we introduce Multilayer Explicit Simulated Synesthesia (MESSy). Instead of enforcing rigid representation matching, which forces the student to distort its native visual features to accommodate complex multimodal topologies, MESSy uses a predictive objective to safely internalize these physical priors. Empirical results demonstrate that our framework significantly outperforms raw data augmentation and traditional distillation. Ultimately, our approach yields highly accurate unimodal students with “synesthetic” latent structures that are inherently aligned with modalities they have never directly observed.
\end{abstract}

\begin{figure}[t]
    \centering
    \includegraphics[width=\linewidth]{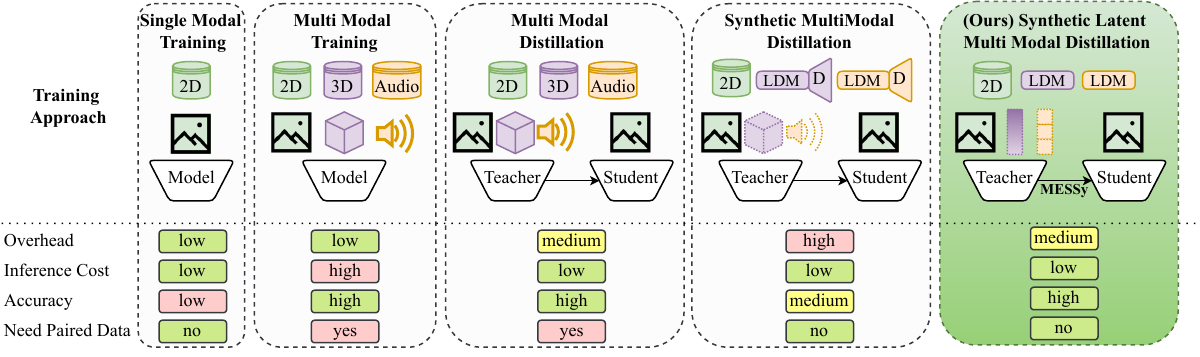}
    \caption{\textbf{Comparison of multimodal learning paradigms.} Traditional multimodal networks achieve high accuracy but require scarce paired datasets and incur prohibitive inference costs. Cross-modal distillation solves the inference bottleneck but remains data-dependent. While naive synthetic distillation removes the need for paired data, it introduces a severe training overhead due to the expensive generative decode-encode loop. Our framework (right) bypasses this bottleneck entirely. By directly distilling undecoded generative latents into a purely visual student, we achieve high multimodal accuracy and efficient unimodal inference without requiring real paired datasets. Our framework reduces 3D generation time by $87.7\%$ and audio VRAM by $23.8\%$ relative to full decoding.}
    \label{fig:abstract_motivation}
\end{figure}  

\section{Introduction}
Human perception is intrinsically multimodal. We understand the physical world not just through visual appearance, but through a continuous integration of sensory signals. The sight of a glass shattering is deeply coupled with its acoustic signature and its geometric fragility. Similarly, neural architectures designed to fuse multiple modalities, such as 2D images with 3D objects and audio, consistently demonstrate superior robustness and accuracy compared to unimodal models. Additional modalities provide indeed complementary physical and semantic priors that help resolve visual ambiguities.

Despite their performance advantages, deploying multimodal networks in real-world applications presents significant challenges. Extracting and processing multiple high-dimensional data streams incurs a substantial computational cost during inference. More practically, auxiliary modalities like structured 3D geometry or clean audio are rarely available in unconstrained test environments. An effective solution to this deployment problem is the paradigm of Learning Using Privileged Information (LUPI)~\cite{vapnik2009lupi} via cross-modal knowledge distillation. By training a heavy multimodal teacher and distilling its knowledge into a lightweight, image-only student, we can achieve high performance while maintaining unimodal inference efficiency. 

However, this approach suffers from a severe data dependency. While image and text pairs can be readily scraped from the web at an unprecedented scale, data for continuous physical modalities such as high-fidelity audio and 3D geometry remains inherently scarce in isolation, and prohibitively rare when strict cross-modal alignment is required.

The rapid advancement of generative artificial intelligence offers a promising mechanism to overcome this data scarcity. Modern cross-modal generators can synthesize highly realistic missing modalities directly from visual inputs or their corresponding text captions~\cite{zhan2023multimodal}. An emerging trend in representation learning exploits these models to perform synthetic multimodal augmentation, creating artificial pairs to train teacher networks~\cite{menadil2023learning,hammoud2024synthclip,brusini2026polygenfullysyntheticvisionlanguage}. While generating synthetic data circumvents the need for real paired datasets, current pipelines overlook a key structural feature of modern generative models. State-of-the-art generators, such as Latent Diffusion Models (LDM), operate within compressed, semantically dense latent spaces. Forcing these latents through an expensive decoding step to produce ``raw'' data creates a highly inefficient \textit{Decode-Encode Loop}. The decoder inevitably injects artifacts and noise irrelevant to classification~\cite{wang2024dealing}, forcing the downstream multimodal classifier to waste parameter capacity re-encoding this noisy data back into a semantic vector.

We propose Direct Latent Augmentation (DLA) to bypass this inefficiency entirely. Instead of fully decoding the synthetic data, we intercept the generative process and extract the undecoded latents. We treat these latents as pristine privileged information. Because they are extracted prior to the generative decoder, they retain maximum semantic density, are free from decoding artifacts, and require a fraction of the computational cost to load and process. We fuse these heterogeneous latents with 2D image patches using a Perceiver-based latent resampler to train our Multimodal Teacher.

Transferring this dense multimodal knowledge to a standard Vision Transformer~\cite{dosovitskiyimage} student presents a secondary challenge. Existing cross-modal distillation, together with standard knowledge distillation loss, minimizes the direct Euclidean distance between the teacher and student embedding spaces. We argue this imposes a rigid geometric isomorphism that causes \textit{capacity interference}: forcing a purely visual student to perfectly mimic a high-dimensional space dedicated to audio and 3D geometry degrades its native visual discriminative ability. To transfer this knowledge effectively, we introduce \textbf{Multilayer Explicit Simulated Synesthesia (MESSy)}. Instead of forcing the student's latent space to identically match the teacher's, MESSy employs lightweight predictive heads. The student learns to predict the teacher's auxiliary modality features from their own visual tokens. This biomimetic approach is inspired by human synesthesia, ensuring the student internalizes the sufficient statistics to recall physical properties without collapsing its native visual manifold. Eventually, we validate our framework on Imagenette, Caltech101, and ImageNet-100. Our contributions are summarized as follows:
\begin{enumerate}
    \item We identify the inefficiencies of the Decode-Encode Loop in generative data augmentation and propose Direct Latent Augmentation (DLA), demonstrating that generative latents provide a cleaner, more efficient supervision signal than raw synthetic data.
    \item We introduce Multilayer Explicit Simulated Synesthesia (MESSy) to solve capacity interference during cross-modal distillation, allowing unimodal students to safely internalize multimodal concepts.
    \item We provide quantitative and qualitative evidence of Artificial Synesthesia, demonstrating that the unimodal student structurally aligns its visual latent space with unobserved physical properties and selectively leverages acoustic and geometric priors to resolve visual ambiguities.
\end{enumerate}

\section{Related Work}
\subsection{Multimodal Classification and LUPI}
Fusing multiple sensory modalities, such as 2D images with audio waveforms or 3D shapes, consistently improves performance on standard classification tasks by leveraging complementary physical priors \cite{nagrani2021attention, kiela2019supervised}. Architectures employing late-fusion, cross-modal attention, or attention bottlenecks \cite{jaegle2021perceiver, xiao2020audiovisual} effectively aggregate these heterogeneous inputs to resolve visual ambiguities. However, deploying such networks in the wild remains largely impractical. Processing high-dimensional, multi-sensory data incurs substantial inference costs and fundamentally relies on the availability of multimodal data at test time, which is frequently absent in real-world applications.

To bridge the gap between the superior accuracy of multimodal training and the efficiency of unimodal inference, researchers rely on the Learning Using Privileged Information (LUPI) paradigm \cite{vapnik2009lupi}. LUPI operates through cross-modal knowledge distillation \cite{hinton2015distilling}, where a multimodal teacher guides a unimodal student. The pioneering work of Gupta et al.~\cite{gupta2016cross} established RGB-to-depth and RGB-to-flow supervision transfer as a cross-modal distillation paradigm, directly inspiring subsequent LUPI pipelines~\cite{hoffman2016cross,garcia2018modality,yuan2018rgb,menadil2023learning}. Despite their success, these methods share a fundamental limitation: they require training datasets of perfectly aligned, real-world multi-sensory triplets. Curating such physically grounded paired data is expensive and difficult to scale, severely limiting the broad application of traditional cross-modal distillation for standard vision tasks.

\subsection{Generative Augmentation and the Decode-Encode Bottleneck}
The rapid maturation of generative AI provides a solution to data scarcity. Generative models are increasingly used for synthetic data augmentation, creating novel views, depth maps, or text captions to enhance unimodal classifiers \cite{tian2024synclr,azizi2023synthetic,he2022synthetic}. 

Closely related to our objective is the recent concept of Learning Using Generated Privileged Information (LUGPI) \cite{menadil2023learning}. LUGPI demonstrates that generating the image modality from text prompts via diffusion and treating it as privileged information improves student classification accuracy. However, their scope remains confined to the highly abundant text and image domains, neglecting the scarce physical modalities, such as 3D structure and audio, that are essential human-like understanding. Crucially, pipelines such as LUGPI operate exclusively in the raw data domain, fully decoding generated signals into human-perceptible formats, like high-resolution images.

Feature-level augmentation within a single visual encoder has also been explored to improve robustness to affine transformations~\cite{sandru2022feature}. In contrast, our DLA framework generates auxiliary-modality latents from frozen cross-modal generators as privileged information, operating across modality boundaries rather than perturbing the model's own feature space.

We argue that this introduces a significant computational and representational inefficiency. Because state-of-the-art generators already perform dense semantic compression in a latent space \cite{ldm,kreuk2022audiogen}, pushing these representations through a final decoder merely to accommodate human perception is counterproductive. This decoding step injects generative artifacts, phase noise, and high-frequency details that act as nuisance variables for classification. Consequently, the downstream multimodal teacher wastes parameter capacity re-encoding this noisy raw data. Our Direct Latent Augmentation (DLA) framework bypasses this decode-encode bottleneck entirely, utilizing the undecoded generative latents directly to provide a cleaner and computationally cheaper supervision signal.

\subsection{Feature-Level Distillation and Capacity Interference}
Knowledge distillation traditionally aligns logits~\cite{hinton2015distilling}; FitNets~\cite{romero2014fitnets} extended this to projector-based hidden-layer alignment, with multi-layer targets stabilizing optimization~\cite{touvron2021training,sun2019patient}. Asymmetric predictor heads, where a student predicts frozen teacher representations, underpin BYOL~\cite{grill2020bootstrap}, data2vec~\cite{baevski2022data2vec,baevski2023efficient}, and their cross-modal extension AV-data2vec~\cite{lian2023avdata2vec}, the closest architectural predecessor to MESSy. JEPA variants~\cite{assran2023ijepa,hu20243djepa} similarly predict abstract teacher embeddings rather than raw pixels. CRD~\cite{tian2020contrastive} shows that relaxing rigid $L_2$ alignment improves cross-modal transfer; Zhao et al.~\cite{zhao2025crossmodal} independently formalize the capacity-interference problem and propose a trainable projector as remedy.

Cross-modal LUPI pipelines standardly penalize the $L_2$ distance between the student's unimodal and the teacher's multimodal embedding~\cite{menadil2023learning,yuan2018rgb,garcia2018modality}, inducing \textit{capacity interference} (Sec.~\ref{sec:stage3}). MESSy adopts the predictor-head pattern of the above methods but targets pooled summaries of \textit{unobserved synthetic auxiliary-modality} tokens, distinct from self-features~\cite{baevski2022data2vec,assran2023ijepa}, the teacher CLS~\cite{menadil2023learning}, or same-modality perturbations~\cite{sandru2022feature} preserving the student's native visual manifold.

\section{Proposed Method} 
\subsection{The Advantage of Latents and Multimodal Synergy}
\label{sec:theory_latents}

Recent advancements in generative AI, particularly Latent Diffusion Models (LDMs) and Flow Matching~\cite{ldm,lipmanflow}, have established a powerful paradigm for high-fidelity data synthesis. These models operate fundamentally in two stages. First, a generative process iteratively transforms initial noise $\epsilon \sim \mathcal{N}(0, \mathbf{I})$ into a structured, semantically dense representation $Z$ within a compressed latent space, guided by a conditioning signal $c$. Second, a pre-trained decoder $\mathcal{D}$ maps this latent vector into the high-dimensional raw data space, producing the human-perceptible signal $\hat{X} = \mathcal{D}(Z)$ (such as an audio waveform or a 3D mesh).

In the context of multimodal data augmentation, the goal is to provide a classifier with auxiliary physical signals that correlate with the target label $Y$. Standard synthetic augmentation pipelines utilize the fully decoded raw data $\hat{X}$. To process this data, the downstream multimodal classifier must employ a modality-specific encoder $\mathcal{E}$, yielding a feature representation $F = \mathcal{E}(\hat{X})$. 

This establishes a deterministic Markov chain from the generated latent to the final classification features: $Z \xrightarrow{\mathcal{D}} \hat{X} \xrightarrow{\mathcal{E}} F$. By the Data Processing Inequality~\cite{dpi}, the mutual information $I$ between the label $Y$ and any representation along this chain cannot increase:
\begin{equation}
I(Y; Z) \geq I(Y; \hat{X}) \geq I(Y; F)
\end{equation}
This inequality exposes a fundamental inefficiency in current synthetic augmentation pipelines. The decoder $\mathcal{D}$ is optimized for human perception, not machine classification. To bridge the gap between the compressed latent space and the raw data manifold, $\mathcal{D}$ injects high-frequency structural details, phase variations, and unavoidable generative artifacts. From the perspective of the classifier, this injected information acts as a noisy channel that degrades the Signal-to-Noise Ratio (SNR) of the underlying semantics. Consequently, the downstream encoder $\mathcal{E}$ must waste parameter capacity attempting to filter out this perceptual noise to recover the semantic concepts already present in $Z$.

Furthermore, our framework allows us to efficiently scale the number of conditioning modalities without amplifying this encoding noise. From an information-theoretic perspective, integrating multiple latent modalities, such as visual ($Z_v$), acoustic ($Z_a$), and geometric ($Z_g$) representations, strictly expands the available semantic signal. By the chain rule of mutual information, the joint information shared with the target label $Y$ is:
\begin{equation}
    I(Y; Z_v, Z_a, Z_g) = I(Y; Z_v) + I(Y; Z_a | Z_v) + I(Y; Z_g | Z_v, Z_a)
\end{equation}
Because conditional mutual information is inherently non-negative:
\begin{equation}
I(Y; Z_v, Z_a, Z_g) \geq I(Y; Z_v)
\end{equation}

Thus, each additional latent space has the potential to provide orthogonal, non-redundant cues, such as structural morphology from 3D latents or unique acoustic signatures, that resolve ambiguities present in the visual signal alone. 

By intercepting the generative pipeline and utilizing the undecoded latents directly as inputs to the multimodal classifier, our framework theoretically guarantees a higher upper bound on $I$. This provides a cleaner, dense supervision signal that maximizes multimodal synergy, while drastically reducing the computational overhead of decoding and re-encoding continuous physical modalities.

\begin{figure}[t]
    \centering
    \includegraphics[width=\linewidth]{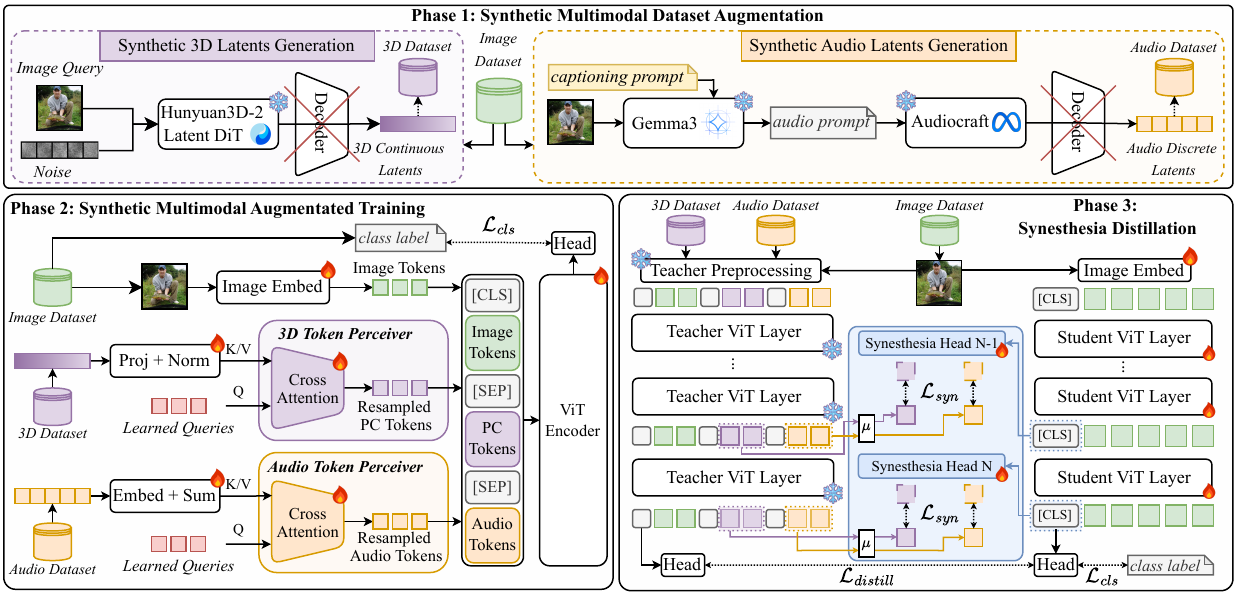}
    \caption{\textbf{Phase 1 (Synthetic Latent Augmentation):} Given an image query, we extract undecoded continuous 3D latents and discrete audio codes directly from generative models, deliberately bypassing the noisy and computationally expensive Decode-Encode Loop. \textbf{Phase 2 (Multimodal Teacher Training):} A Multimodal Teacher is trained to classify these augmented triplets. We fuse standard image tokens with the heterogeneous generative latents using modality-specific Perceiver cross-attention blocks. \textbf{Phase 3 (MESSy Distillation):} We transfer this multimodal knowledge to a lightweight, purely visual student. Using our MESSy objective, lightweight auxiliary heads predict the teacher's multimodal features across multiple intermediate layers. This allows the student to learn the physical priors avoiding capacity interference.}
    \label{fig:GA}
    \vspace{-1em}
\end{figure}  

\subsection{Stage 1: Synthetic Latent Augmentation}
\label{sec:stage1}

The first stage of our framework constructs an offline, multimodal latent dataset from a unimodal image corpus (Phase 1, \cref{fig:GA}). For a given RGB image query $X_v$, we extract semantically corresponding generative latents for 3D geometry and audio without ever synthesizing the final raw signals. 

For the 3D modality, we employ the pre-trained Hunyuan3D-2 Latent DiT~\cite{hunyuan3d22025tencent} to obtain the $X_v$ image-conditioned continuous 3D latents, $Z_{3D} \in \mathbb{R}^{N_{3D} \times D_{3D}}$. Crucially, the 3D latent-to-mesh decoder is bypassed, saving significant computational time and preserving semantic density.

For the audio modality, direct image-to-audio generative models remain scarce and lack the robust, zero-shot generalization capabilities of their text-conditioned counterparts. To bridge this modality gap, we employ Gemma3~\cite{gemma3} to process the image query and generate a descriptive captioning prompt of the likely acoustic events in the scene. To strictly prevent label leakage, this prompting process is carefully designed: the model is never provided with the ground-truth class name, and we implement an automatic filtering mechanism that regenerates the text if the generated caption inadvertently contains the class name or any of its WordNet synonyms (full prompt and pipeline details are provided in the supplementary material). We then input this audio prompt into Audiocraft~\cite{audiocraft}. Because audio generators operate using discrete latent spaces quantized by neural audio codecs, we intercept the generation process immediately after the autoregressive transformer samples the acoustic codebook, extracting the undecoded Audio Discrete Latents $Z_{aud} \in \mathbb{Z}^{N_{aud}}$. We deliberately extract and save the undecoded discrete latent codes $Z_{aud}$ rather than the generator's internal continuous embeddings. This not only maximizes offline storage and I/O efficiency, but also allows our downstream Teacher network to learn a task-specific embedding lookup that prioritizes semantic classification over acoustic waveform reconstruction.

Both $Z_{3D}$ and $Z_{aud}$ are saved to disk, creating a highly compressed, information-dense multimodal dataset while completely avoiding the Decode-Encode Loop. We report full per-modality generation details in the supplementary material.

\subsection{Stage 2: Synthetic Multimodal Augmented Training}
\label{sec:stage2}

In Phase 2, we train a multimodal teacher network, $\mathcal{T}$, to classify the augmented triplets $(X_v, Z_{3D}, Z_{aud})$. Fusing these modalities presents a dimensional challenge: the visual input consists of standard spatial patches, the 3D input consists of continuous latent sequences of variable length, and the audio input consists of discrete codebook tokens.

To project these heterogeneous inputs into a unified representation space, we design a modality-agnostic fusion architecture utilizing Perceiver-inspired~\cite{jaegle2021perceiver,alayrac2022flamingo} cross-attention blocks. The visual image $X_v$ is processed through a standard patch embedding layer to produce Image Tokens $E_v$. For the 3D modality, the continuous latents $Z_{3D}$ are projected and normalized, then passed as Keys and Values to a 3D Token Perceiver. By cross-attending with a fixed number of Learned Queries, the Perceiver compresses the variable-length latents into a fixed budget of Resampled 3D Tokens $E_{3D}$. 

Similarly, the discrete audio codes $Z_{aud}$ are passed through a learned embedding lookup, summed, and fed as Keys and Values to an Audio Token Perceiver. This outputs a fixed budget of Resampled Audio Tokens $E_{aud}$.

The final input sequence to the Teacher's ViT Encoder is constructed by concatenating the modality tokens, explicitly bounded by learnable separator tokens ($[\mathtt{SEP}]$), alongside a prepended classification token ($[\mathtt{CLS}]$):
\begin{equation}
    E_{input} = \left[ [\mathtt{CLS}], E_v, [\mathtt{SEP}_1], E_{3D}, [\mathtt{SEP}_2], E_{aud} \right]
\end{equation}
The Multimodal Teacher $\mathcal{T}$ processes this joint sequence and is optimized using a standard cross-entropy loss, $\mathcal{L}_{\mathrm{cls}}$, against the ground-truth class labels.

\subsection{Stage 3: Distillation via Multilayer Explicit Simulated Synesthesia (MESSy)}
\label{sec:stage3}
In the final phase, our objective is to transfer the robust multimodal representations of the frozen teacher $\mathcal{T}$ into a lightweight, unimodal student network $\mathcal{S}$, which receives only the standard RGB images as input. Along with standard logits distillation, recent works using LUPI and LUGPI frameworks achieve this by an additional loss term enforcing a strict geometric isomorphism, minimizing the direct $L_2$ distance between the student's visual embedding and the teacher's multimodal embedding. However, we argue this is fundamentally flawed for asymmetric cross-modal distillation. The teacher's embedding resides in a joint manifold shaped by visual, acoustic, and geometric features, whereas the student is constrained by the information capacity of a purely visual input. Forcing a lower-capacity unimodal embedding to perfectly mimic this high-dimensional topology induces \textit{capacity interference}: a phenomenon where the network struggles to simultaneously optimize for fundamentally conflicting representations, leading to destructive gradient conflict \cite{wang2020makes}. The student is forced to distort its optimal visual decision boundaries to geometrically accommodate the missing modalities, which actively degrades its primary discriminative capabilities.

To resolve this capacity bottleneck, we propose Multilayer Explicit Simulated Synesthesia (MESSy), illustrated in Phase 3 of Figure~\ref{fig:GA}. Rather than enforcing rigid $L_2$ spatial equality $||e^S - e^T||_2^2$ between the visual student embedding $e^S$ and the multimodal teacher embedding $e^T$, MESSy requires the student to encode only the sufficient statistics necessary to reconstruct the auxiliary modalities, thereby inducing ``Simulated Synesthesia''. To ensure computational efficiency and avoid the overhead of sequence-to-sequence prediction, we do not force the student to reconstruct the full sequence of synthetic tokens. Instead, let $e_{aud}^{T, l}$ and $e_{3D}^{T, l}$ represent the global, average-pooled summary vectors of the acoustic and geometric tokens extracted from the teacher at layer $l$. We attach lightweight, non-linear projection heads $H_{aud}$ and $H_{3D}$ (implemented as 2-layer MLPs) to the student's corresponding classification tokens $e_{\text{CLS}}^{S, l}$. The MESSy objective is defined as the MSE over the deepest $K$ layers of the Vision Transformer:
\begin{equation}
\mathcal{L}_{\mathrm{MESSy}} = \sum_{l=L-K}^{L} \mathbb{E} \left[ || H_{aud}(e_{\text{CLS}}^{S, l}) - e_{aud}^{T, l} ||_2^2 + || H_{3D}(e_{\text{CLS}}^{S, l}) - e_{3D}^{T, l} ||_2^2 \right]
\end{equation}

From an Information Bottleneck perspective, this predictive objective maximizes the mutual information $I(e_{\text{CLS}}^S; e_{aud}^T, e_{3D}^T)$ by utilizing the projection heads to bridge the spatial gap between the manifolds, safely bypassing capacity collapse. We apply this to multiple layers following recent advancements in transformer distillation showing that multi-layer supervision stabilizes the optimization trajectory of self-attention blocks \cite{touvron2021training, sun2019patient}. By applying MESSy across the final $K$ layers, we provide useful gradient signals without interfering with early-stage visual feature extraction (we ablate this number in the supplementary material).

During inference, the auxiliary heads $H$ are entirely discarded, leaving the student with an enriched visual latent space and zero additional computational overhead. The total training objective combines standard cross-entropy ($\mathcal{L}_{\mathrm{cls}}$) against the hard labels, classical logit-level knowledge distillation ($\mathcal{L}_{\mathrm{distill}}$), and our explicit synesthesia loss ($\mathcal{L}_{\mathrm{MESSy}}$):
\begin{equation}
\mathcal{L}_{\mathrm{Total}} = \alpha \mathcal{L}_{\mathrm{cls}} + (1 - \alpha) \mathcal{L}_{\mathrm{distill}} + \beta \mathcal{L}_{\mathrm{MESSy}}
\end{equation}

\section{Experiments}
\subsection{Experimental Setup}
\label{sec:setup}

\textbf{Datasets.} We evaluate on Imagenette2-320~\cite{Howard_Imagenette_2019}, Caltech101~\cite{caltech101}, and ImageNet-100~\cite{deng2009imagenet,in100}. Focusing on these benchmarks allows us to rigorously ablate modality combinations and perform expensive raw-vs-latent generative comparisons within a feasible computational budget. To verify that our method scales beyond these benchmarks, we also evaluate our best DLA+MESSy configuration on the full ImageNet-1K dataset~\cite{deng2009imagenet} (Sec.~\ref{sec:imagenet1k}). We execute 3 independent runs per experiment, reporting averages and standard deviations ($\pm$) in tables, and averages in plots for clarity.

\textbf{Generative Pipeline.} For 3D latent extraction, we utilize the pre-trained \texttt{Hunyuan3D-2-mini} Latent Diffusion Model (FP16 variant). We extract the continuous image-conditioned spatial latents directly from the DiT flow-matching pipeline, bypassing the final mesh decoder. For audio latent extraction, we first prompt the \texttt{Gemma-3-12B-IT} large language model with the visual image to generate a short, descriptive acoustic caption. This caption conditions \texttt{AudioGen-Medium}. We intercept the generation process at the autoregressive transformer output, extracting the discrete tokens from the 4 parallel acoustic codebooks prior to EnCodec waveform decoding.

\textbf{Architecture and Training.} Both our multimodal teacher ($\mathcal{T}$) and unimodal student ($\mathcal{S}$) are based on the standard Vision Transformer (ViT-B/16) architecture, ensuring fair capacity comparisons. 
For the Teacher, we introduce modality-specific Latent Resamplers using Perceiver-style cross-attention with 8 heads. These resamplers compress the variable-length generative latents into 100 tokens for 3D geometry and 100 tokens for audio. During Teacher training, we apply an aggressive modality dropout rate of $p=0.8$ to the acoustic and geometric tokens. We find this to be crucial to prevent the model from over-relying on the synthetic modalities (detailed in the supplementary material). 

\textbf{Distillation Details.} The Teacher is frozen after 50 epochs of training. The Student is then trained from scratch for 50 epochs using AdamW~\cite{adamw} optimization and a cosine annealing learning rate scheduler. For the distillation objective, we set the standard Knowledge Distillation temperature to $T=2.0$ and the Cross-Entropy weight to $\alpha=0.25$. For the MESSy objective, we extract features from the last $K=3$ layers (ablated in the supplementary material) and we set the loss weight to $\beta=1.0$.

\subsection{Multimodal Teacher Performance: The Latent Advantage}
\label{sec:teacher_results}
Before evaluating downstream accuracy, we quantify the computational savings of this bypass during the offline dataset augmentation phase. Bypassing the expensive mesh decoder for 3D generation reduces inference time by $87.7\%$ and saves over $600$\,MB of peak VRAM. Similarly, extracting discrete audio tokens prior to waveform decoding yields a $20.6\%$ reduction in generation time and cuts peak VRAM usage by $23.8\%$ (nearly $2.8$\,GB). Full benchmarking details are provided in the supplementary material.
\begin{figure*}[t]
    % --- Left side: The Plot ---
    \begin{minipage}[c]{0.50\textwidth}
        \centering
        \includegraphics[width=\linewidth]{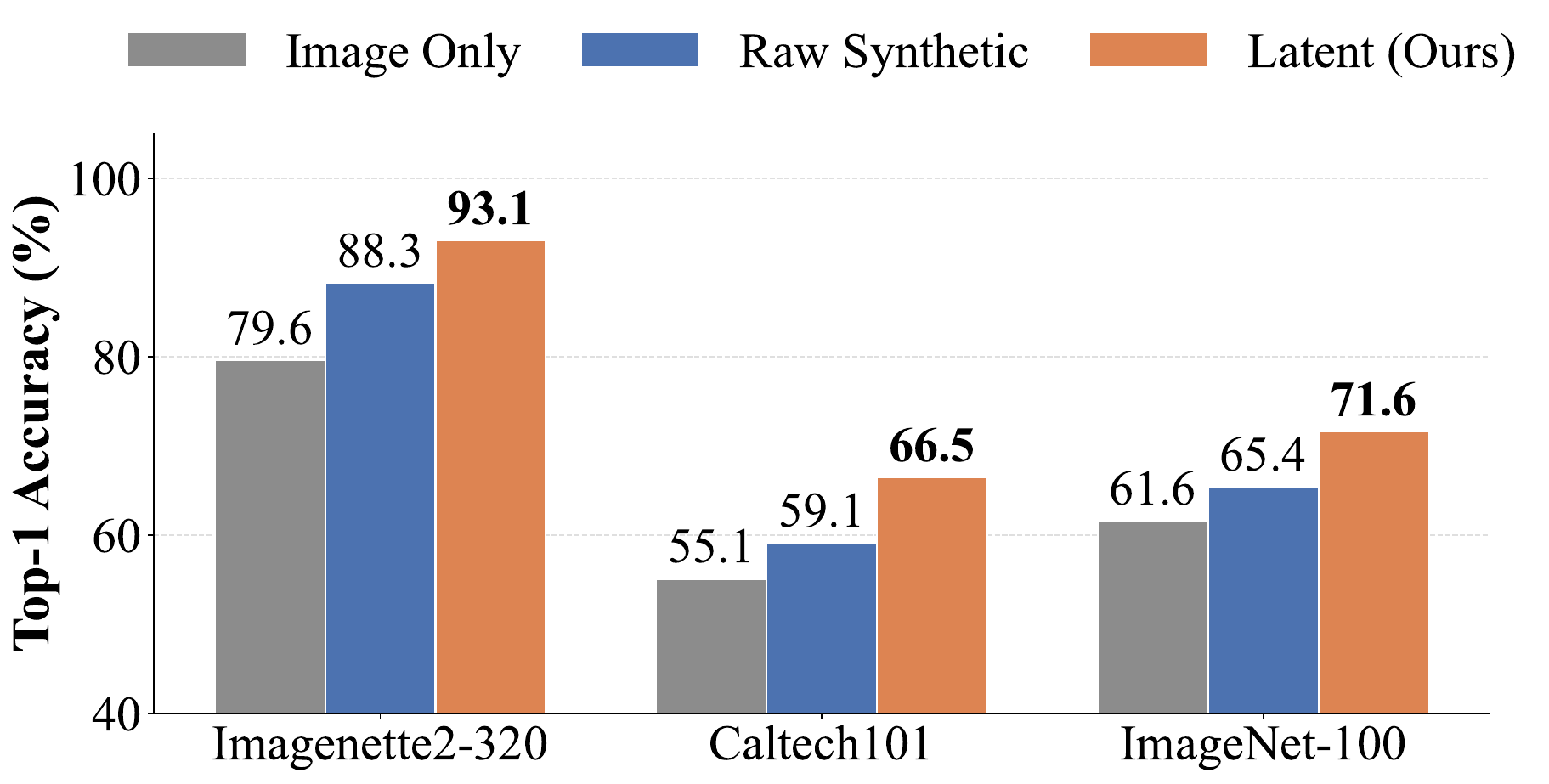}
        \caption{Representation efficiency of the Multimodal Teacher. DLA consistently outperforms fully decoded raw synthetic data across all benchmarks.}
        \label{fig:raw_vs_latents}
    \end{minipage}\hfill
    % --- Right side: The Table ---
    \begin{minipage}[c]{0.48\textwidth}
        \centering
        \captionof{table}{Multimodal Teacher validation accuracies. We compare the Image-Only baseline, fully decoded Raw synthetic data, Raw data with the same Perceiver resampler architecture, and our undecoded Generative Latents.}
        \label{tab:raw_vs_latents_table}
        \resizebox{\textwidth}{!}{% Resizes table to fit minipage perfectly
        \begin{tabular}{lccc}
        \toprule
        \textbf{Method} & \textbf{Imagenette2} & \textbf{Caltech101} & \textbf{ImageNet-100} \\
        \midrule
        Image Only & $79.63\err{0.32}$ & $55.10\err{0.27}$ & $61.57\err{0.53}$ \\
        Raw       & $88.28\err{0.65}$ & $59.08\err{1.96}$ & $65.40\err{0.90}$ \\
        Raw+Perc. & $88.92\err{0.43}$ & $60.21\err{1.12}$ & $66.13\err{0.77}$ \\
        Latents   & $\mathbf{93.10}\err{0.19}$ & $\mathbf{66.47}\err{1.42}$ & $\mathbf{71.59}\err{0.45}$ \\
        \bottomrule
        \end{tabular}
        }
    \end{minipage}
\end{figure*}

\Cref{fig:raw_vs_latents} and \cref{tab:raw_vs_latents_table} compare multimodal teachers accuracies when augmented with raw vs latent synthetic data. For the raw synthetic data encoding we use a PointNet++ style local feature tokenizer, while we convert WAV to Mel-spectrograms and then embed them via 2D convolution. Training the Teacher directly on generative latents significantly outperforms training on fully decoded raw synthetic data across all benchmarks. For instance, on Caltech101, the latent teacher achieves an absolute improvement of $+7.39\%$ over the raw synthetic teacher and $+11.37\%$ over the Image-only baseline, with similarly substantial gains observed on ImageNet-100 and Imagenette2-320. These results empirically validate our analysis (Sec. \ref{sec:theory_latents}): generative decoding injects perceptual noise that degrades the semantic Signal-to-Noise Ratio, whereas undecoded latents provide dense, highly discriminative privileged information. \begin{wrapfigure}{R}{0.55\textwidth}
    \centering
    \includegraphics[width=\linewidth]{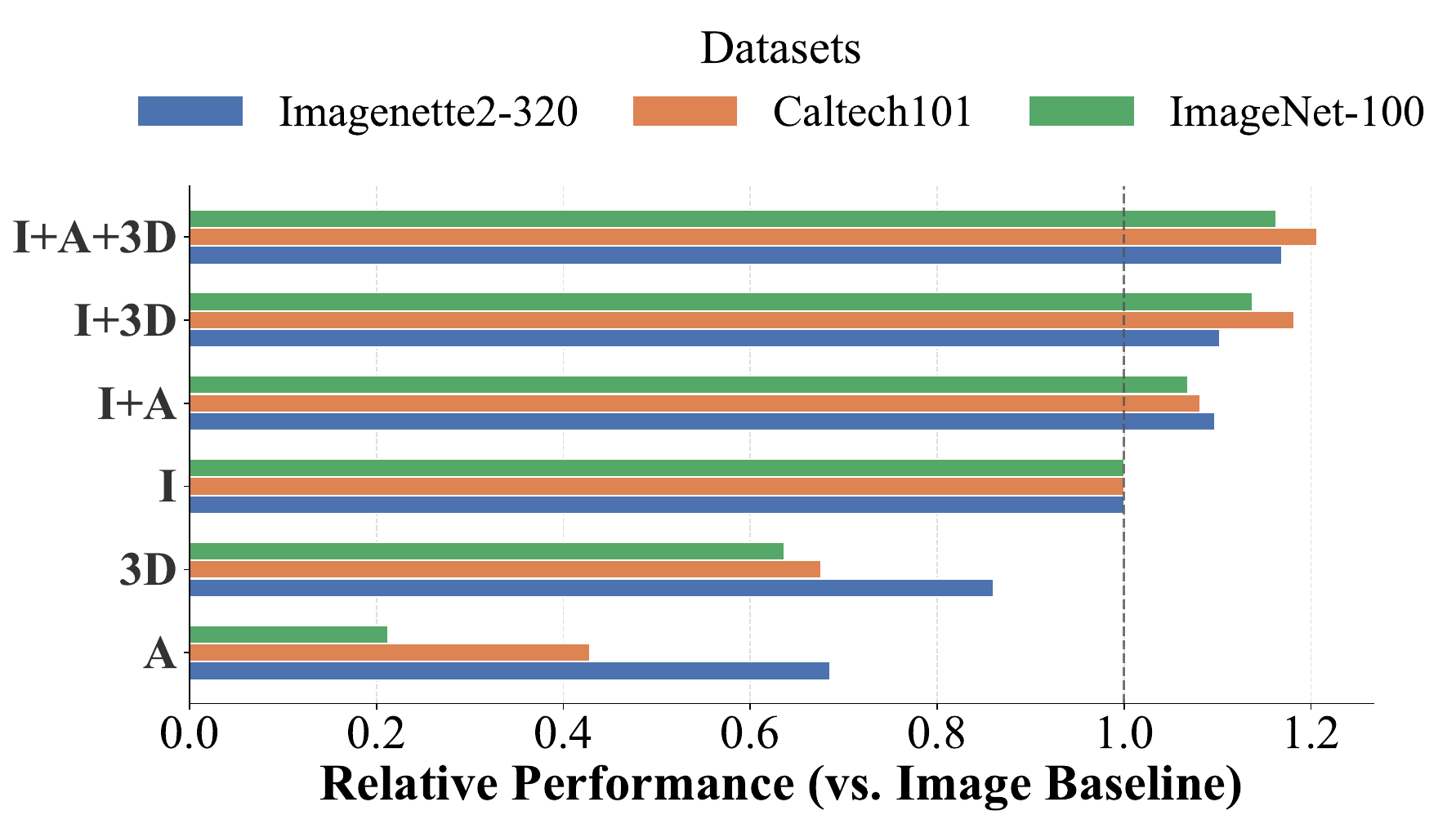}
    \caption{Modality contribution analysis.}
    \label{fig:modalities_contributions}
\end{wrapfigure} To rule out that this gain arises from a difference in encoder architecture rather than representation quality, we also train a Raw teacher that feeds decoded modalities through the same Perceiver resampler used for latents (\textit{Raw+Perc.} in \cref{tab:raw_vs_latents_table}). The Perceiver resampler accounts for at most $1.13$ pp ($\approx$14\% of the gap on average), confirming that the dominant factor is the undecoded representation itself, in line with the Data Processing Inequality.

Furthermore, we perform a study on the relative contribution of each modality \cref{fig:modalities_contributions,tab:modality_comparison}. We observe that 3D geometry and audio provide orthogonal, complementary supervision. On ImageNet-100, the Image+3D ($70.03\%$) and Image+Audio ($65.76\%$) combinations both independently improve upon the Image-Only baseline ($61.57\%$). Fusing all modalities (the ``Full'' configuration) consistently yields the highest upper bound performance ($71.59\%$). 

Crucially, we evaluate the networks trained solely on the isolated synthetic modalities to ensure the model is not simply exploiting a generative shortcut. We find that the synthetic modalities perform poorly on their own, falling far below the real Image-Only baseline. The Audio-Only baseline achieves $13.07\%$ on ImageNet-100, well above the $1\%$ random chance for 100 classes, which confirms that the acoustic latents do carry class-relevant signal, yet far below the $61.57\%$ Image-Only accuracy, ruling out any trivial label leakage from the generation pipeline. The 3D-Only baseline ($39.20\%$) is stronger, reflecting the richer spatial structure encoded by the Hunyuan3D latents. Both modalities therefore act as complementary physical priors that sharpen the visual encoder rather than as shortcuts, validating the integrity of our label-leakage prevention pipeline.

\begin{table}[t]
    \centering
    \scriptsize
    \caption{Performance comparison across different modality combinations. We report the average validation accuracy and standard deviation ($\pm$) across 3 runs, alongside the absolute performance delta ($\Delta$) relative to the Image-Only baseline.}
    \label{tab:modality_comparison}
    \begin{tabular}{l|cc|cc|cc}
        \toprule
        \multirow{2}{*}{\textbf{Modalities}} & \multicolumn{2}{c|}{\textbf{Imagenette2}} & \multicolumn{2}{c|}{\textbf{Caltech101}} & \multicolumn{2}{c}{\textbf{ImageNet-100}} \\
        & Accuracy & $\Delta$ & Accuracy & $\Delta$ & Accuracy & $\Delta$ \\
        \midrule
        Image Only            & $79.63\err{0.32}$ & $-$ & $55.10\err{0.27}$ & $-$ & $61.57\err{0.53}$ & $-$ \\
        Audio Only            & $54.61\err{0.85}$ & $-25.02$ & $23.61\err{1.10}$ & $-31.49$ & $13.07\err{0.54}$ & $-48.50$ \\
        3D Only               & $68.53\err{0.43}$ & $-11.10$ & $37.23\err{2.84}$ & $-17.87$ & $39.20\err{0.73}$ & $-22.37$ \\
        \midrule
        Image + Audio         & $87.37\err{0.24}$ & $+7.74$  & $59.58\err{0.83}$ & $+4.48$  & $65.76\err{0.04}$ & $+4.19$ \\
        Image + 3D            & $87.80\err{0.15}$ & $+8.17$  & $65.15\err{0.24}$ & $+10.05$ & $70.03\err{0.53}$ & $+8.46$  \\
        Image + 3D + Audio    & $\mathbf{93.10}\err{0.19}$ & $\mathbf{+13.47}$ & $\mathbf{66.47}\err{1.42}$ & $\mathbf{+11.37}$ & $\mathbf{71.59}\err{0.45}$ & $\mathbf{+10.02}$ \\
        \bottomrule
    \end{tabular}
\end{table}

\subsection{Synesthesia Distillation Efficacy}
\label{sec:student_results}
\begin{figure*}[t]
    \centering
    \begin{minipage}[c]{0.52\textwidth}
        \centering
        % Replace 'distillation_comp.pdf' with your actual file path
        \includegraphics[width=\textwidth]{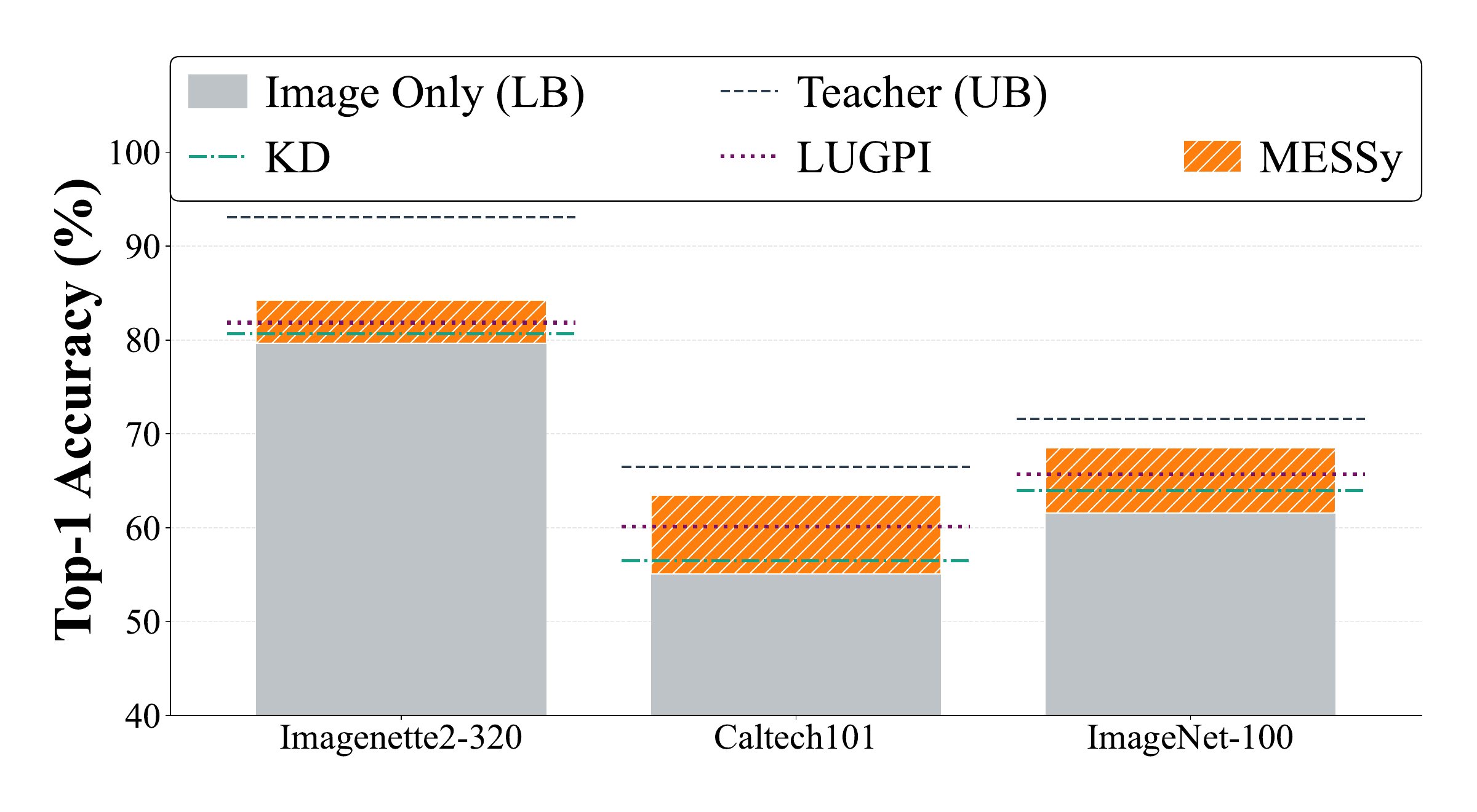}
        \caption{Our approach (MESSy) against the RGB-only Lower Bound, KD, LUGPI, and the Synthetic Upper Bound (Teacher UB).}
        \label{fig:distillation}
    \end{minipage}\hfill
    \begin{minipage}[c]{0.46\textwidth}
        \centering
        \resizebox{\textwidth}{!}{%
        \begin{tabular}{lccc}
            \toprule
            \textbf{Method} & \textbf{Imagenette2} & \textbf{Caltech101} & \textbf{ImageNet-100} \\
            \midrule
            Image & $79.63\err{0.32}$ & $55.10\err{0.27}$ & $61.57\err{0.53}$ \\
            Synth. UB & $93.10\err{0.19}$ & $66.47\err{1.42}$ & $71.59\err{0.45}$ \\
            \midrule
            KD & $80.66\err{0.20}$ & $56.49\err{0.42}$ & $63.97\err{0.59}$ \\
            LUGPI & $81.85\err{0.51}$ & $60.12\err{0.76}$ & $65.70\err{0.68}$ \\
            \textbf{MESSy} & $\mathbf{84.23}\err{0.40}$ & $\mathbf{63.52}\err{0.91}$ & $\mathbf{68.54}\err{0.83}$ \\
            \bottomrule
        \end{tabular}%
        }
        % Using the caption package allows \captionof for tables inside figure environments
        \makeatletter\def\@captype{table}\makeatother
        \caption{Top-1 Accuracy (\%) comparison of unimodal distillation paradigms. \textit{Synth.\ UB} is the DLA latent teacher trained on RGB + synthetic modalities. All student models evaluate \textbf{purely on visual inputs} at inference time.}
        \label{tab:distillation_results}
    \end{minipage}
\end{figure*}

Having established a robust multimodal upper bound, we evaluate the efficacy of transferring this knowledge to a unimodal, image-only Student ($\mathcal{S}$) (\cref{fig:distillation,tab:distillation_results}). Our MESSy Student achieves the best unimodal inference performance, outperforming the baseline by $+8.42\%$ on Caltech101 and $+6.97\%$ on ImageNet-100, effectively bridging the gap to the synthetic upper bound.

We compare MESSy against KD and LUGPI, all using the same DLA latent teacher (LUGPI applies $L_2$ matching to our synthetic representations, not a raw-data teacher). KD yields only marginal gains: logit-level matching cannot capture the geometric and acoustic semantics embedded in the teacher's intermediate representations. LUGPI improves substantially but is limited by the rigid $L_2$ alignment it imposes; MESSy's predictive heads relax this constraint, adding $+3.40\%$ over LUGPI on Caltech101 and $+2.84\%$ on ImageNet-100 (Sec.~\ref{sec:stage3}).

\noindent\textbf{Capacity interference evidence.} We freeze both IN-100 students and evaluate them on a CIFAR-100 linear probe, a purely visual task seen by neither student during training. LUGPI scores $42.4\%$ vs.\ MESSy's $48.1\%$ ($+5.7$ pp). Since both students share the same teacher and training images, the gap can only be attributed to the distillation objective: $L_2$ matching degrades visual discriminability to accommodate the multimodal teacher topology, while MESSy's predictive heads preserve it.

\noindent\textbf{MESSy target ablation and DLA isolation.} \cref{tab:ablation_messy_dla} ablates two factors on ImageNet-100. A CLS-Pred variant uses the same MLP heads but targets teacher CLS tokens: switching from $L_2$ to a projection head adds $+1.34$ pp; switching the prediction target to auxiliary-modality summaries adds $+1.50$ pp, showing the gain is not merely architectural. Replacing the DLA teacher with a raw-data teacher costs $3.39$ pp under LUGPI and $4.51$ pp under MESSy, indicating the latent-space advantage carries through to the student.

% --- Side-by-Side Figure and Table ---

\subsection{Emergent Synesthesia}
\label{sec:nmi_results}

To confirm emergent synesthesia, the student's visual latent space must reflect the topology of unobserved modalities. We cluster 3D and audio independently into ground-truth topological spaces and measure their NMI against the model's visual embedding clusters. Two feature types per modality span different abstraction levels: \textit{Hand-Crafted (HC)}, covering low-level acoustics (MFCCs~\cite{mfcc}, spectral centroids) and covariance-based geometry; and \textit{Deep}, using CLAP~\cite{clap} for audio and PointNet~\cite{qi2017pointnet} for 3D. Cluster counts scale with class vocabulary (50 for Imagenette2, 150 otherwise; details in supplementary).

As shown in \cref{fig:synesthesiaAlignment}, the baseline aligns poorly with both modality spaces. LUGPI improves alignment on the Deep semantic dimension (e.g.\ Audio Deep NMI $+0.111$ on ImageNet-100) but not on the physical HC space ($+0.026$): $L_2$ matching transfers high-level semantics but cannot imprint low-level physical topology. MESSy yields larger and more consistent gains across both dimensions (Audio Deep $+0.154$, Audio HC $+0.135$). The HC improvement is particularly telling: these features encode raw acoustic statistics and geometric covariance with no semantic content, so the alignment cannot arise from label co-occurrence and is evidence of genuine structural synesthesia.

\begin{figure*}[t]
    \centering
    \includegraphics[width=\linewidth]{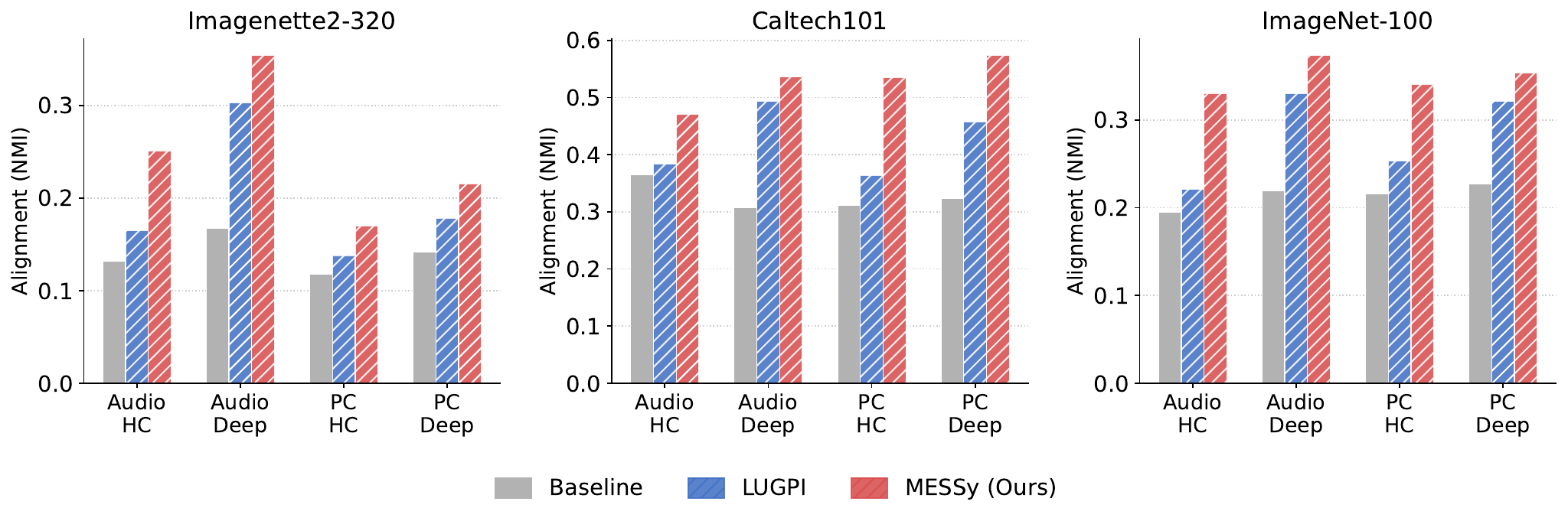}
    \vspace{-2em}
    \caption{Emergent alignment between the models' visual latent spaces and the unseen modalities. We compare the Image-Only baseline, the LUGPI-distilled student, and our MESSy student across both low-level physical (HC) and semantic (Deep) feature clusters. LUGPI improves Deep semantic alignment but barely affects low-level HC features. MESSy yields consistently larger and more balanced gains across both spaces.}
    \label{fig:synesthesiaAlignment}
\end{figure*}

\begin{figure*}[t]
    \centering
    \includegraphics[width=\linewidth]{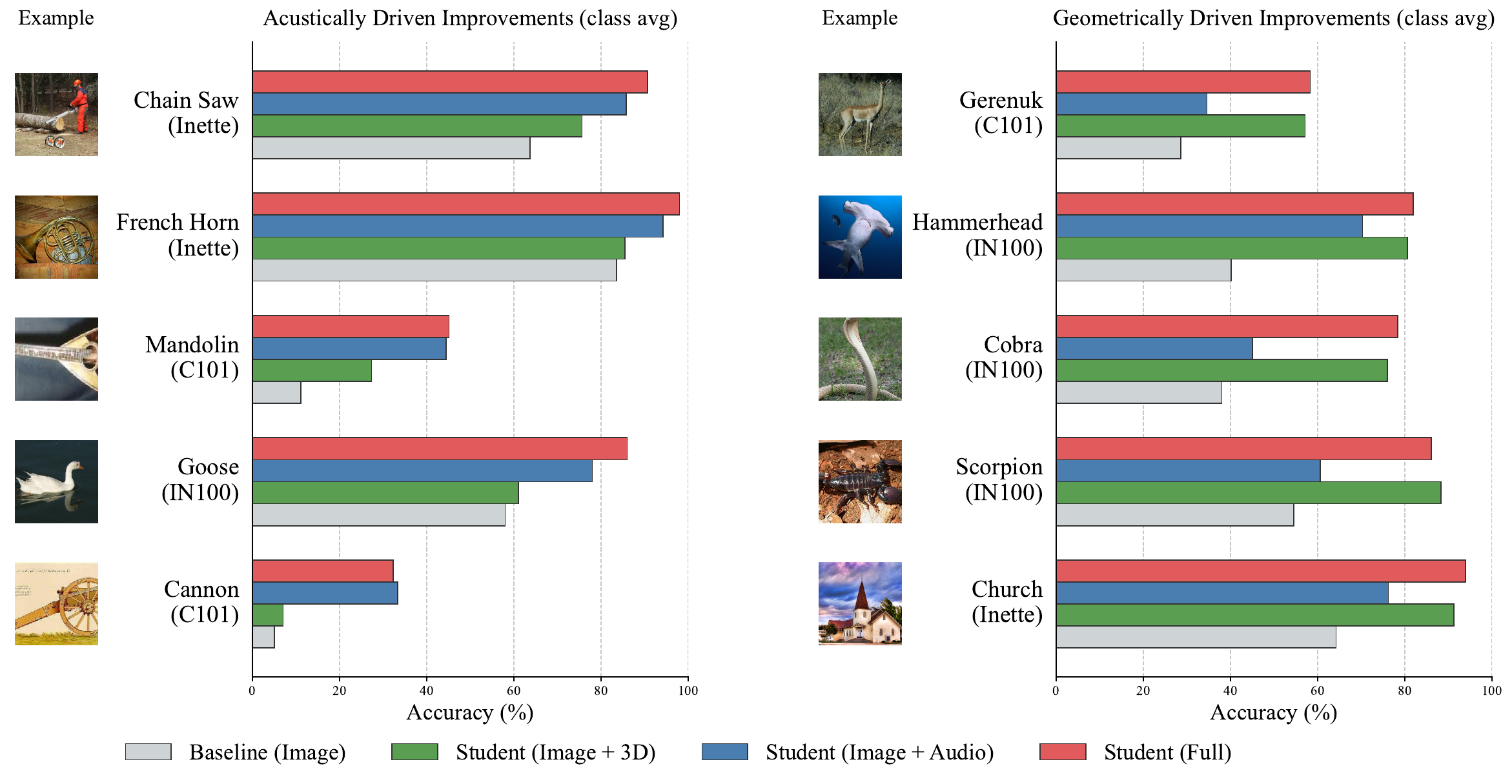}
    \vspace{-2em}
    \caption{Qualitative analysis of class-level accuracy improvements driven by explicitly simulated synesthesia. We compare the Image-Only baseline against students distilled from bimodal teachers (Image+3D and Image+Audio) and the Full teacher (Image+Audio+3D). \textbf{Left:} Classes where the simulated acoustic prior drives the most significant gains, typically characterized by distinct sound profiles (e.g., instruments). \textbf{Right:} Classes where the geometric prior dominates the improvement, predominantly consisting of species and objects with complex, distinctive 3D morphological traits.}
    \label{fig:qualitative_improvements}
\end{figure*}

\subsection{Qualitative Analysis: Disentangling Modality-Driven Priors}
\label{sec:qualitative_analysis}
To understand the mechanisms driving our performance gains, we perform a qualitative class-level analysis. We compare the Image baseline against students distilled from bimodal and Full teachers. While the Full student achieves the highest overall accuracy, analyzing the bimodal models isolates the impact of specific physical priors (Fig.~\ref{fig:qualitative_improvements}).

\noindent\textbf{Acoustically Driven Improvements.} 
The left column of Fig.~\ref{fig:qualitative_improvements} shows classes where the Image+Audio student yields the largest gains. These classes (e.g., \textit{Chain Saw}, \textit{French Horn}, \textit{Mandolin}) are defined by highly distinct acoustic signatures. By distilling acoustic latents via MESSy, the visual encoder leverages simulated sound profiles to resolve visual ambiguities, such as distinguishing a mandolin from similarly shaped stringed instruments.

\noindent\textbf{Geometrically Driven Improvements.} 
The right column highlights classes benefiting most from the Image+3D student, predominantly animals with unique morphological traits (e.g., the \textit{Gerenuk}'s long neck or the \textit{Hammerhead}'s cephalofoil). Distillation from 3D latents forces the student to internalize a robust spatial prior, making its vision-only representations significantly more effective.

These results empirically validate Simulated Synesthesia: the unimodal student successfully internalizes intuitive acoustic and geometric priors that align with human semantic understanding.

\subsection{Scalability to ImageNet-1K}
\label{sec:imagenet1k}
\vspace{-0.2em}
To verify that DLA+MESSy scales beyond the smaller benchmarks, we run the same pipeline on the full ImageNet-1K (ILSVRC-2012)~\cite{deng2009imagenet}, processing all 1.28M training images. The DLA teacher reaches $71.32\%$ ($+7.15\%$ over image-only), and the MESSy student reaches $69.05\%$ ($+4.88\%$) evaluating on RGB only. The student recovers roughly $68\%$ of the teacher gap, consistent with results on ImageNet-100, suggesting the framework scales predictably with dataset size.

\begin{table}[t]
    \centering
    \begin{minipage}[t]{0.65\linewidth}
        \centering
        \scriptsize
        \caption{Ablations on ImageNet-100. \textit{Left:} distillation objective. \textit{Right:} DLA vs.\ raw teacher per objective.}
        \label{tab:ablation_messy_dla}
        \vspace{-0.5em}
        \begin{tabular}{lc|lcc}
            \toprule
            \textbf{Objective} & \textbf{Top-1} & \textbf{Teacher} & \textbf{Obj.} & \textbf{Top-1} \\
            \midrule
            LUGPI     & $65.70\pm0.68$ & Raw & LUGPI & ${\sim}62.31$ \\
            CLS-Pred       & $67.04\pm0.61$ & Raw & MESSy & ${\sim}64.03$ \\
            \textbf{MESSy} & $\mathbf{68.54\pm0.83}$ & DLA & LUGPI & $65.70\pm0.68$ \\
                           &                & DLA & \textbf{MESSy} & $\mathbf{68.54\pm0.83}$ \\
            \bottomrule
        \end{tabular}
    \end{minipage}%
    \hfill
    \begin{minipage}[t]{0.31\linewidth}
        \centering
        \scriptsize
        \caption{Scalability on ImageNet-1K.}
        \label{tab:imagenet1k_results}
        \vspace{-0.5em}
        \begin{tabular}{lcc}
            \toprule
            \textbf{Method} & \textbf{Top-1} & \textbf{$\Delta$} \\
            \midrule
            Image-Only & $64.17$ & $-$ \\
            \midrule
            Synth.\ UB & $71.32$ & $+7.15$ \\
            \midrule
            \textbf{MESSy} & $\mathbf{69.05}$ & $\mathbf{+4.88}$ \\
            \bottomrule
        \end{tabular}
    \end{minipage}
\end{table}

\section{Conclusion}
We introduced DLA to bypass the inefficient decode-encode loop of generative multimodal training, utilizing pristine latents directly as privileged information. To seamlessly transfer this dense knowledge, our MESSy distillation objective allows a purely visual student to internalize acoustic and geometric priors without suffering from capacity interference. While our current framework relies on frozen, off-the-shelf generators, limiting task-specific latent co-adaptation, the empirical benefits are significant. Ultimately, our approach demonstrates that unimodal networks can achieve artificial synesthesia, successfully hallucinating multimodal properties to resolve visual ambiguities, likewise to human inherent multimodal understanding and reasoning.

\section*{Acknowledgments}
 This paper is supported by the PNRR-PE-AI FAIR project funded by the NextGeneration EU program. This work was partially financially supported by JST ASPIRE Program, Japan, Grant Number JPMJAP2303. This work was partially supported by the JSPS Postdoctoral Fellowship for Research in Japan (Fellowship ID: P24752).

\bibliographystyle{splncs04}
\bibliography{main}
\end{document}